\documentclass[10pt,twocolumn,letterpaper]{article}

\usepackage[pagenumbers]{cvpr} 

\usepackage{graphicx}
\usepackage{amsmath}
\usepackage{amssymb}
\usepackage{booktabs}

\usepackage{url}
\usepackage[labelformat=simple]{subcaption}

\newcommand{\tabscaleA}{0.8}

\newcommand{\R}{\mathbb{R}}

\usepackage[pagebackref,breaklinks,colorlinks]{hyperref}

\usepackage[capitalize]{cleveref}
\crefname{section}{Sec.}{Secs.}
\Crefname{section}{Section}{Sections}
\Crefname{table}{Table}{Tables}
\crefname{table}{Tab.}{Tabs.}

\def\cvprPaperID{*****} 
\def\confName{CVPR}
\def\confYear{2022}

\begin{document}

\title{Object-ABN: Learning to Generate Sharp Attention Maps\\ for Action Recognition}

\author{
Tomoya Nitta\\
Nagoya Institute of Technology\\
Nagoya, Japan
\and
Tsubasa Hirakawa, Hironobu Fujiyoshi\\
Chubu University,
Aichi, Japan
\and
Toru Tamaki\\
Nagoya Institute of Technology\\
Nagoya, Japan
}

\maketitle

\begin{abstract}
In this paper we propose an extension of the Attention Branch Network (ABN) by using instance segmentation for generating sharper attention maps for action recognition. Methods for visual explanation such as Grad-CAM usually generate blurry maps which are not intuitive for humans to understand, particularly in recognizing actions of people in videos. Our proposed method, Object-ABN, tackles this issue by introducing a new mask loss that makes the generated attention maps close to the instance segmentation result. Further the PC loss and multiple attention maps are introduced to enhance the sharpness of the maps and improve the performance of classification. Experimental results with UCF101 and SSv2 shows that the generated maps by the proposed method are much clearer qualitatively and quantitatively than those of the original ABN.
\end{abstract}

\section{Introduction}

Action recognition \cite{DBLP:journals/corr/abs-2012-06567,Video_Action_Understanding_2021,Video_Transformers_Survey,10.3389/frobt.2015.00028,DBLP:journals/corr/abs-1806-11230}
is one of long-standing topics in computer vision
and still actively studied thanks to the emergence of deep learning techniques and large datasets.
The task is to classify a trimmed video clip (typically several second-long)
into pre-defined action categories \cite{Feichtenhofer_2020_CVPR,Feichtenhofer_2019_ICCV}.
It is a basis of other video-related tasks such as
temporal action localization \cite{Xia_2020_IEEEaccess} where
untrimmed videos are divided into short clips
whose features are extracted by using action recognition models.

Deep models, not limited to action recognition but also other tasks, are difficult to investigate
because of its black-box nature, hence
visual explanations
have been studied in the field of explainable AI
\cite{xai_survey_2018_IEEEaccess,xai_survey_2020,HUANG2020100270,xai_medical_survey_2021,xai_survey_2021_entropy,xai_survey_2022}.
These are attempts to generate saliency maps
to visualize which of the parts in the scene are important for classification,
and many methods have been utilized, including
Grad-CAM \cite{Selvaraju_2017_ICCV}, 
Score-CAM \cite{Wang_2020_CVPR_Workshops}, LRP \cite{LRP_2015_PlosOne}, and ABN \cite{Fukui_2019_CVPR}.
A common problem of these approaches is that generated maps are often blurry and ambiguous,
hence difficult to understand and interpret for human observers \cite{xai_survey_2018_IEEEaccess,hiley2019explainable}.
Recent studies have therefore been attempting to improve the quality and sharpness of the maps by focusing on objects,
for example, by combining Grad-CAM and LRP \cite{Relevance-CAM_CVPR2021},
applying LRP to Vision Transformers \cite{Chefer_2021_CVPR},
improving ABN with Score-CAM \cite{Lee_2021_ICCV},
and even human intervention \cite{Mitsuhara2021} or additional supervision \cite{Li_2018_CVPR}.

For action recognition, 
this blurry map issue still remains
while many visual explanation methods tailored for videos have been proposed.
The challenge is to make the model focus on the regions of people who perform the actions in the scene.
To this end, some works evaluate visualization results 
by checking the peak of the map being inside bounding boxes of humans (called pointing games)
\cite{DBLP:journals/ijcv/ZhangBLBSS18,DBLP:conf/eccv/ZhangLBSS16,Perturbation_WACV2021,Zhenqiang_2022TCSVT,EB-RNN_CVPR2018}.
However, this is not a direct approach to the blurry map issue,
and models still suffer from from the representation bias;
models may use clues of backgrounds of the scene instead of the foreground
\cite{DBLP:journals/corr/abs-2012-06567,DBLP:journals/corr/HeSSK16,Li_2018_ECCV}.

\begin{figure}[t]
    \centering

    \begin{minipage}{.8\linewidth}
        \includegraphics[width=\linewidth]{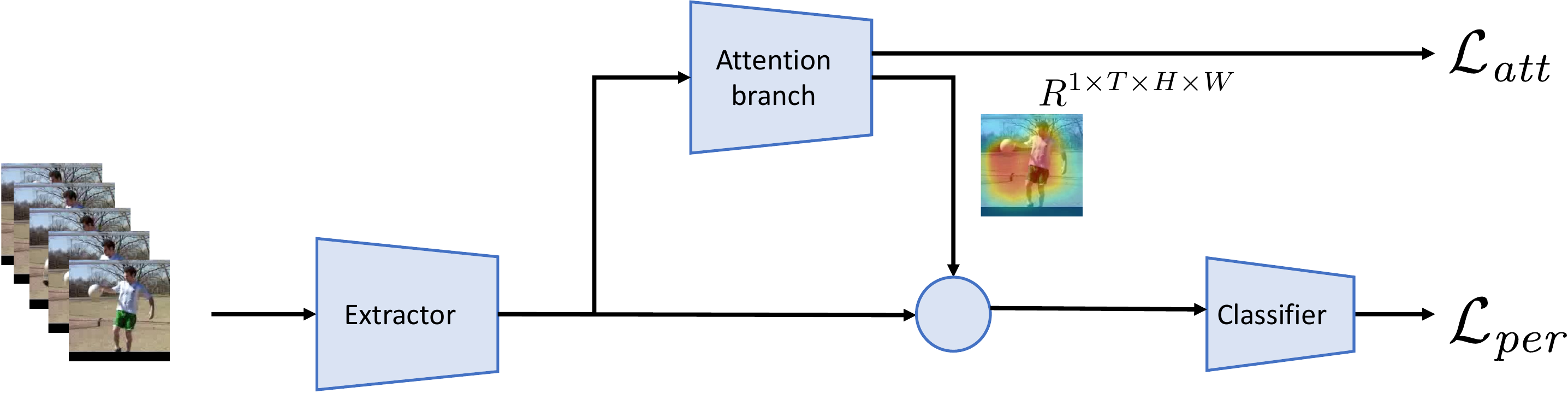}
        \subcaption{}
        \label{fig:ABN}
    \end{minipage}

    \vspace{1em}
    
    \begin{minipage}{.8\linewidth}
        \includegraphics[width=\linewidth]{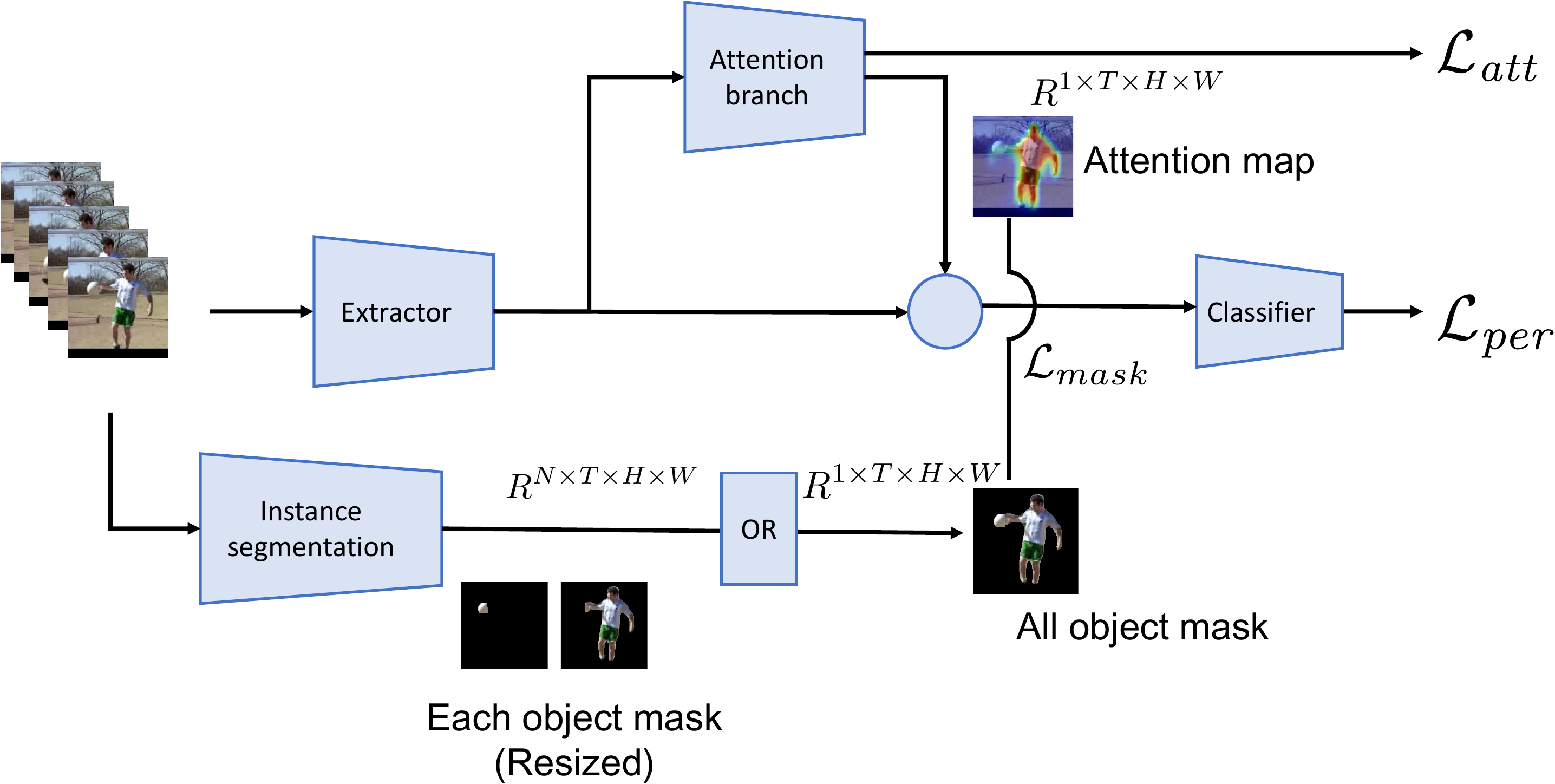}
        \subcaption{}
        \label{fig:Object-ABN}
    \end{minipage}

    \caption{Overview of models of (a) ABN for action recognition, and (b) the proposed Object-ABN.}
    \label{fig:ABN_proposed_method}
\end{figure}

In this paper, we propose \emph{Object-ABN}, a direct and simple approach to the blurry map issue of action recognition.
The key idea is to combine an off-the-shelf instance segmentation model
with Attention Branch Network (ABN) \cite{Fukui_2019_CVPR} (Fig.\ref{fig:ABN}).
ABN is a classification model with attention and perception branches;
the attention branch generates an attention map and uses it as weights of the feature map fed to the perception branch.
The attention branch has its own classifier to improve the predictive power of the attention map.
In the proposed method (Fig.\ref{fig:Object-ABN}),
a constraint is added so that the attention map is close to the instance segmentation result.
This is expected to lead to a better explanability with a sharper map
focusing on people and objects in the scene.
However, a possible drawback is the performance-explainability trade-off
when adding explanation modules to a model
\cite{xai_survey_2018_IEEEaccess,xai_survey_2020,xai_survey_2021_entropy,xai_survey_2022}.
To mitigate this, we propose to use Prototype Conformity (PC) loss \cite{Mustafa_2019_ICCV}
that enforces features to be separated in several clusters.
In the following sections, we summarize related works,
and we briefly describe ABN, then explain the proposed Object-ABN in detail.
Then we show experimental results with two datasets, UCF101 \cite{DBLP:journals/corr/abs-1212-0402} and Something-Something v2 \cite{Goyal_2017_ICCV},
with quantitative evaluation on the sharpness of attention maps.

\section{Related works}

Explainable AI (XAI)
\cite{xai_survey_2018_IEEEaccess,xai_survey_2020,HUANG2020100270,xai_medical_survey_2021,xai_survey_2021_entropy,xai_survey_2022}
has become an important topic particularly for deep learning models.
Visual explanation (or visual attribution) is a topic of XAI,
which is to generate a map (sometimes called saliency map, attention map, or attribution map)
that visually indicates where in the image the model is focusing on for classification.

Methods are often categorized into post-hoc and intrinsic \cite{xai_survey_2018_IEEEaccess,xai_survey_2020,HUANG2020100270,xai_medical_survey_2021,xai_survey_2021_entropy,xai_survey_2022}.
\emph{Post-hoc} methods are used to analyse a single prediction of a trained model,
and the name comes from the fact that the visual explanation is done after the model has been trained and fixed.
This category includes well-known methods such as 
CAM \cite{Zhou_2016_CVPR}, Grad-CAM \cite{Selvaraju_2017_ICCV}, and LRP \cite{LRP_2015_PlosOne}.
These were originally proposed for images, but can be used for videos as well,
so they have been used as a baseline for comparison.
In addition, some works proposed post-hoc methods tailored for videos by extending methods for images.
For example,
DevNet \cite{DevNet_CVPR2015} used gradient-based Deep Inside CNN \cite{Simonyan14a_deep_inside} and graph-cut for extracting important regions,
EB-RNN \cite{EB-RNN_CVPR2018} is based on 
Excitation Backprop (EB) \cite{DBLP:conf/eccv/ZhangLBSS16,DBLP:journals/ijcv/ZhangBLBSS18} for models wit CNN and RNN,
and EP-3D and ST-EP \cite{Perturbation_WACV2021,Zhenqiang_2022TCSVT} extends
Extremal Perturbation (EP) \cite{Fong_2019_ICCV} for spatio-temporal 3D volumes.
LRP/DTD \cite{LRP_2015_PlosOne} has also been applied to videos \cite{DBLP:series/lncs/AndersMSM19,DBLP:journals/corr/abs-1908-01536}.
Few methods have been proposed specific for video;
saliency tubes \cite{Saliency_Tubes_ICIP2019} proposed an additional module for visualizing spatio-temporal tubes,
and class feature pyramids \cite{Class_Feature_Pyramids_ICCVW2019} proposed a feature back-propagation of 3D CNN.

The post-hoc approach is useful for investigating the behavior of a given model, particularly sensitively can be visualized by showing maps for each category.
However, post-hoc methods based on gradients (Grad-CAM \cite{Selvaraju_2017_ICCV}) and back-propagation (LRP \cite{LRP_2015_PlosOne})
are inherently difficult to generate sharp maps because class-prediction information flows from the top to the bottom though the network.
Therefore some attempts have been proposed to make the map sharper;
for example, Relevance-CAM \cite{Relevance-CAM_CVPR2021} combines LRP and Grad-CAM.
Perturbation-based methods \cite{Fong_2019_ICCV,DBLP:journals/ijcv/ZhangBLBSS18} suffer from the same problem, as well as a high computation cost for perturbing masks many times for video volumes.

\emph{Intrinsic} methods has its own mechanism of visual explanation in the model itself. This approach has an advantage that the model is designed to have explanability in the first place \cite{xai_survey_2022}, and that visual explanation during a training phase would be useful for practitioners to check the model performance qualitatively.
Because the explanation mechanism of an intrinsic method is a part of the model,
there are a great variety of model architectures.
Sharma \etal \cite{DBLP:journals/corr/SharmaKS15,sharma2016actrecICLR}
used LSTM to predict the soft attention map of the next frame,
which were later extended to video captioning with attention \cite{pmlr-v37-xuc15}.
Attention pooling \cite{NIPS2017_67c6a1e7} decomposed a 3D attention map with 2nd order pooling and rank-1 approximation.
Interpretable spatio-temporal attention 
\cite{Meng_2019_ICCV_Workshops}
used spatial and temporal attention via ConvLSTM.
Recent self-attention mechanisms are also introduced
in STA-TSN \cite{STA-TSN_PLOSONE2022} and GTA \cite{he2021gta},
as well as Transformer-based video models \cite{Video_Transformers_Survey}.
Some of these methods do not aim to visual explanation,
and the blurry map issue still remains for videos because
the ability of temporal modeling, which is useful for classification,
may be harmful to capture sharp spatial attention maps.

In this paper, we focus on ABN \cite{Fukui_2019_CVPR}, an intrinsic method proposed for images.
ABN first extracts features, then the attention branch computes an attention map
which is multiplied to the feature map, then the perception branch classifies the weighted feature map (see Fig.\ref{fig:ABN}).
The attention map of ABN is useful as visual explanation because the attention map directly specifies the importance of the feature map that is used for classification.
However, sometimes the attention map of ABN differs greatly from the human intuition.
To alleviate this problem,
a Human-in-the-loop (HITL) framework \cite{Iwayoshi2021} was proposed
to enable human operators to modify the attention map of ABN.
This results in a sharper attention map that are easy to interpret by humans,
leading to a better explanability through attention visualization.
However, human intervention on videos that requires frame-by-frame annotations is costly and impractical.
In contrast, our proposed Object-ABN scales to a large amount of video frames
because it is trained in an end-to-end manner by introducing
instance segmentation as an additional self-supervision.

\section{Method}

In this section, we describe ABN \cite{Fukui_2019_CVPR}
for action recognition. Although the original ABN was proposed for classifying images,
notations are aligned with the the proposed models described below.

\subsection{ABN}

ABN consists of feature extractor $E$, attention branch $A$, and perception branch $P$.
Let input video clip be
$x \in \R^{T_\mathrm{in} \times 3 \times H_\mathrm{in} \times W_\mathrm{in}}$,
where $T_\mathrm{in}$ is the number of video frames,
$H_\mathrm{in}, W_\mathrm{in}$ are height and width of the frame.
The corresponding ground-truth action label is denoted by $y \in \{0, 1\}^{L}$, where $L$ is the number of categories.

First, the extractor takes a video clip and output feature maps
$h_1 = E(x) \in \R^{T \times C \times H \times W}$,
where $C$ is the channel size.
Then the attention branch takes it and
generates frame-wise (unconstrained) attention maps $M_u \in [0, 1]^{T \times 1 \times H \times W}$ and 
class prediction $y_a \in [0, 1]^{L}$ as well.
The maps $M_u$ are applied to feature maps $h_1$ of each frame separately
to generate $h_2 \in \R^{T \times C \times H \times W}$ as
\begin{align}
    h_{2,t,c} &= h_{1,t,c} M_{u,t},
    \text{ for } t=1,\ldots,T \ c=1,\ldots,C.
\end{align}
The loss attached to the attention branch is
$
\mathcal{L}_\mathit{att} = \mathcal{L}_\mathit{CE} (y_a, y),
$
where $\mathcal{L}_\mathit{CE}$ is a cross entropy loss.

The perception branch $P$ takes weighted feature maps $h_2$ and
outputs prediction $y_p = P(h_2) \in [0, 1]^{L}$.
The loss of this branch is
$
    \mathcal{L}_\mathit{per} = \mathcal{L}_\mathit{CE} (y_p, y).
$

\subsection{Object-ABN}

As mentioned before,
the attention maps $M_u$ generated by ABN is
blurry and often different from areas where people consider important.
In this study, we assume that regions of people and objects in the scene are important for 
identifying action categories,
and we enforce on the shape of the attention map being closer to the scene objects.
We call this Object-ABN.

To this end, we propose to apply a pre-trained instance segmentation model
to each frame of the video clip, which generates the ground-truth object masks
$M_\mathit{gt} \in \{0, 1\}^{T \times N(t) \times H \times W}$
for instances that appear in the video clip.
$N(t)$ is the number of instances detected at frame $t$, so it differs at different frames.
We aggregate the object masks to a single channel mask
$M'_\mathit{gt} \in \{0, 1\}^{T \times 1 \times H \times W}$
by using logical OR as follows;
\begin{align}
    M'_{\mathit{gt},t} &= \bigcup_{c=1}^{N(t)} M_{\mathit{gt},t,c}.
\end{align}

This is used to compute the following mask loss
\begin{align}
    \mathcal{L}_\mathit{mask} &= \mathcal{L}_\mathrm{MSE}(M_o, M_\mathit{gt}'),
    \label{eq:mask_loss}
\end{align}
which is a mean squared error (MSE) loss between the ground-truth
object masks and the attention maps $M_o$.
Here we use $M_o$ as the ABN attention maps (instead of $M_u$)
to show that it is constrained by the object masks $M_\mathit{gt}'$.

\subsection{Using multiple attention maps}

Object-ABN generates the object-constrained attention maps $M_o$ for each frame $t$,
that is, $M_{o,t}$ for $t=1,\dots,T$. Each map has a single channel, which means that
the same attention map is applied to all $C$ channels of features $h_1$.
However, different channels of $h_1$ may capture different concepts of the scene,
and it might be desirable to use different attention maps for different channels.

Therefore, in this study, we propose to use multiple attention maps by using multi-head attention (MHA).
Specifically, we use $C'$ heads to output maps in the attention branch, and 
each head $c'$ generates attention maps $M_{c'} \in [0, 1]^{T \times 1 \times H \times W}$.

The number of heads $C'$ can be different from the number of channels $C$,
and we align the dimensions as follows.
First, we apply the attention map $M_{c'}$ to each channel $c$ of $h_1$;
\begin{align}
    \begin{split}
        h^{c'}_{1,t,c} =& \ 
        h_{1,t,c} M_{c',t}
        \in \R^{T \times C \times H \times W},
    \end{split}
\end{align}
then concatenate them in the channel direction;
\begin{align}
    h_1' &= \mathrm{cat}(h_1^{1}, h_1^{2}, \ldots, h_1^{C'})
    \in \R^{T \times (CC') \times H \times W},
\end{align}
and use a $1 \times 1$ convolution
\begin{align}
    h_2 &= \mathrm{conv}(h_1') \in \R^{T \times C \times H \times W},
\end{align}
with the kernel size of $1 \times (CC') \times 1 \times 1$
to generate $h_2$ with the appropriate dimension.

In this study, we set $C'=3$.
This means there are three attention maps $M_1, M_2$, and $M_3$,
and we denote them as $M_u, M_o$, and $M_b$, respectively.
$M_u$ is the unconstrained attention maps as in the original ABN,
and $M_o$ is the object-constrained map $M_o$ of Object-ABN.
$M_b$ is the attention maps of the background.
In action recognition, it is known that the background can be a clue for classification 
\cite{DBLP:journals/corr/abs-2012-06567,DBLP:journals/corr/HeSSK16}
because of the representation bias \cite{Li_2018_ECCV}.
We use two maps for foreground and background by explicitly separating them.

We introduce the following loss for three attention maps;
\begin{align}
    \mathcal{L}_\mathit{mha} =& \
    \mathcal{L}_\mathrm{MSE}(M_o, M_\mathit{gt}')  + 
    \lambda_\mathit{b} \mathcal{L}_\mathrm{MSE}(M_b, 1 - M_\mathit{gt}'),
\end{align}
where $\lambda_\mathit{b}$ is a weight.
The first term is the same with the mask loss \eqref{eq:mask_loss}.
The second term is for the background attention maps $M_b$
and uses the inverse of the ground-truth object masks.
Note that we don't use any losses for $M_u$,
and let the network to obtain the map by itself
because the unconstrained maps might be useful like as in the original ABN.

\subsection{PC Loss}

When creating attention maps,
it would be desirable to have features well separated in the middle of the network,
particularly in the attention branch,
because the attention branch can generated maps suitable for each action categories.
To this end,
we introduce the Prototype Conformity (PC) loss \cite{Mustafa_2019_ICCV},
which encourages cluster to be generated in the latent space and facilitates feature separation.
The use of clustered features would be advantageous for generating sharp attention maps while preserving accuracy.

We use the PC loss for features $f \in R^{d}$ in the attention branch.
The loss is represented by
\begin{align}
\begin{split}
  \mathcal{L}_{PC} =& \
   \lambda_{PC_1} \| f - w^c_y \|_2 \\
  &
  -\frac{\lambda_{PC_2}}{K - 1}\sum_{j \ne y} \left( \| f - w^c_j \|_2 +  \| w^c_y - w^c_j \|_2 \right),
  \end{split}
\end{align}
where $y$ is the label, 
$K$ is the number of clusters,
and $w^c_j$ is the $j$-th trainable centroid.

The total loss is one of the following;
\begin{align}
    \mathcal{L} =& 
    \mathcal{L}_\mathit{per} 
    + \lambda \mathcal{L}_\mathit{att}
    + \lambda_\mathit{mask} \mathcal{L}_\mathit{mask}
    + \lambda_\mathit{PC} \mathcal{L}_\mathit{PC} \\
    \mathcal{L} =& 
    \mathcal{L}_\mathit{per} 
    + \lambda \mathcal{L}_\mathit{att}
    + \lambda_\mathit{mha} \mathcal{L}_\mathit{mha}
    + \lambda_\mathit{PC} \mathcal{L}_\mathit{PC},
\end{align}
where $\lambda, \lambda_\mathit{mask}, \lambda_\mathit{mha}$, and
$\lambda_\mathit{PC}$ are weights.

\section{Experiment}

\subsection{Datasets}

We used two datasets in the experiments; UCF101 \cite{DBLP:journals/corr/abs-1212-0402}, and
something-something v2 (SSv2) \cite{Goyal_2017_ICCV}.

UCF101 \cite{DBLP:journals/corr/abs-1212-0402} has 101 classes of human actions,
consisting of a training set of about 9500 videos and a validation set of about 3500 videos.
Each video was collected from Youtube,
with an average length of 7.21 seconds.
We report the performance of the first split.

SSv2 \cite{Goyal_2017_ICCV} consists of a training set of 168913 videos,
a validation set of 24777 videos.
Each video is about 2 to 6 second-long (average 4.03 seconds), filmed by a crowd worker.
The video contains 173 different templates as action categories,
such as ``Dropping [something] into [something]'' that represents the action performed on objects.

\subsection{Experimental setting}

\noindent\textbf{Training.}\ 
From a video in the training set, we sampled 16 frames with a stride of four frames
(starting at randomly decided frame) to make an input clip.
We resized the shorter side of the frame randomly in the range of 256 to 320 pixels while keeping the aspect ratio, 
randomly cropped a square of size $224 \times 224$ pixels,
and then performed the horizontal flip with a probability of 50\%.
The optimizer used for training was Adam \cite{adam} with the learning rate of $10^{-4}$,
and the number of training epochs was set to 50.

\noindent\textbf{Validation.}\ 
We sampled one clip is sampled from a video in the validation set as in the training,
and resized the short side to 256 pixels while maintaining the aspect ratio,
then cropped the square patch of size $224\times224$ pixels in the center of the frame.

\noindent\textbf{Evaluating sharpness.}\ 
For a quantitative evaluation of the sharpness of attention maps, we propose to use entropy of the maps.
If the attention maps are blurry across the entire frames,
the distribution of values of the maps becomes broad and the entropy increases.
If the attention maps are sharp, the distribution should be polarized toward 0 and 1,
entropy should decrease, and the boundary between people and background is expected to be sharper.
Therefore, we use the entropy as an indicator of the sharpness the attention maps.
However, the entropy decreases when the attention maps are flat and values falls within a certain range.
To mitigate this,
we normalize the attention map at each frame so that the minimum and maximum values of maps are 0 and 1, respectively.

To compute the entropy,
we create a histogram of attention maps with $N=10$ bins from 0 to 1.
The frequency $\mathit{hist}[i]$ of each bin $i$ of the histogram is the normalized discrete probability $p_i$,
which is used to compute the entropy as follows;
\begin{align}
    \mathrm{entropy} &= \sum_{i=1}^N - p_i \log_{2} p_i, \quad
    p_i = \frac{\mathit{hist}[i]}{\sum_{j=1}^N \mathit{hist}[j]}.
\end{align}
The entropy is calculated for each frame of the video clip,
and the entropy of the video is calculated by averaging the entropy of all frames.

As a reference, the maximum of the entropy is achieved when values are $p_i = 1/N$.
Since $N=10$ in our case, $\log_{2}10 \approx 3.332$ is the maximum value.

\noindent\textbf{Model.}\ 
As a backbone of ABN,
we used X3D-M \cite{Feichtenhofer_2020_CVPR} pre-trained on Kinetics400 \cite{kay2017kinetics}.
We divided the X3D-M model in two between the third and fourth ResBlocks,
using the first half as the feature extractor and the second half as the perception branch.
We added an attention branch comprising of three ResBlocks and two conv layers, and the features immediately after the three ResBlocks were used to compute the PC loss.

The resulting model takes an input video clip of size
$T_\mathrm{in} \times H_\mathrm{in} \times W_\mathrm{in} = 16 \times 244 \times 244$,
and generates an attention map of size $T \times H \times W = 16 \times 14 \times 14$
which is the spatial resolution of the third ResBlock of the X3D-M.

\noindent\textbf{Parameters.}\ 
The parameters used in the experiment were set as follows;
$\lambda = 1,
\lambda_\mathit{b} = 1,
\lambda_\mathit{mask} = 10,
\lambda_\mathit{mha} = 10,
\lambda_\mathit{PC} = 10^{-4},
\lambda_\mathit{PC_1} = 1$, and
$\lambda_\mathit{PC_2} = 10^{-3}$.
The number of clusters $K$ was set to $L$, the number of categories.

\begin{table}[t]
    \centering
    \caption{
        The performances and entropy values
        with different configurations for the validation set of UCF101.
    }
    \label{tab:1}

    \scalebox{\tabscaleA}{
        \begin{tabular}{c@{\ }c@{\ }c@{\ }c@{\ }|@{\hspace{.5em}}c@{\hspace{.5em}}c@{\hspace{.5em}}c@{\hspace{.5em}}c@{\hspace{.5em}}c}
            &&&&& & entropy & entropy & entropy\\
             $\mathcal{L}_\mathit{per/attn}$ & $\mathcal{L}_\mathit{mask}$ & $\mathcal{L}_\mathit{mha}$ & 
             $\mathcal{L}_\mathit{PC}$ & top-1 & top-5 & $M_u$ & $M_o$ & $M_b$ \\
             \hline
             \checkmark & & & & 93.96 & 99.15 & 3.064\\
             \checkmark &  & & \checkmark & 94.68 & 99.47 & 3.041\\
             \checkmark & \checkmark &  & & 93.62 & 99.26 & & 2.026 & \\
             \checkmark & \checkmark & & \checkmark & 93.80 & 98.97 & & 1.469 &  \\
             \checkmark & & \checkmark & & 88.93 & 98.04 &2.850 & 1.360 & 1.356\\
             \checkmark & & \checkmark & \checkmark & 87.76 & 97.27 & 2.815 & 1.388 & 1.414 \\ \hline
             \multicolumn{4}{l@{\ }|@{\hspace{.5em}}}{interpretable attention \cite{Meng_2019_ICCV_Workshops} } & 87.11 \\
             \multicolumn{4}{l@{\ }|@{\hspace{.5em}}}{STA-TSN \cite{STA-TSN_PLOSONE2022} RGB} & 83.4\phantom{0} \\
             \multicolumn{4}{l@{\ }|@{\hspace{.5em}}}{STA-TSN \cite{STA-TSN_PLOSONE2022} RGB+flow} & 92.8\phantom{0} \\
             \multicolumn{4}{l@{\ }|@{\hspace{.5em}}}{ST-SAWVLAD \cite{DBLP:journals/ieicetd/ChengXMLGY21}} & 94.8\phantom{0} \\
        \end{tabular}
        }
\end{table}

\begin{figure}[t]
    \centering

    \begin{minipage}{0.9\linewidth}
    \centering
    \includegraphics[width=\linewidth]{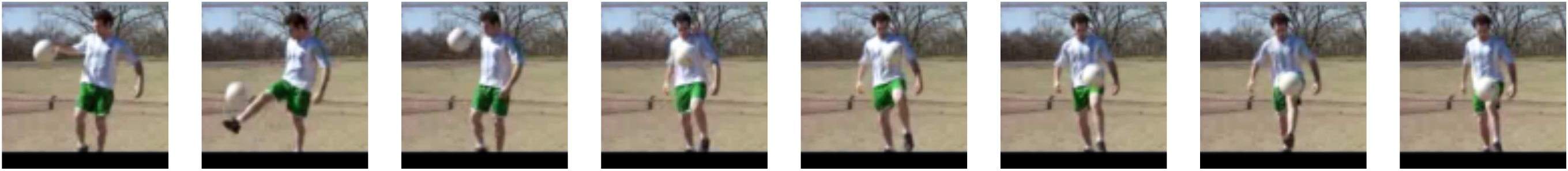}\\
    \includegraphics[width=\linewidth]{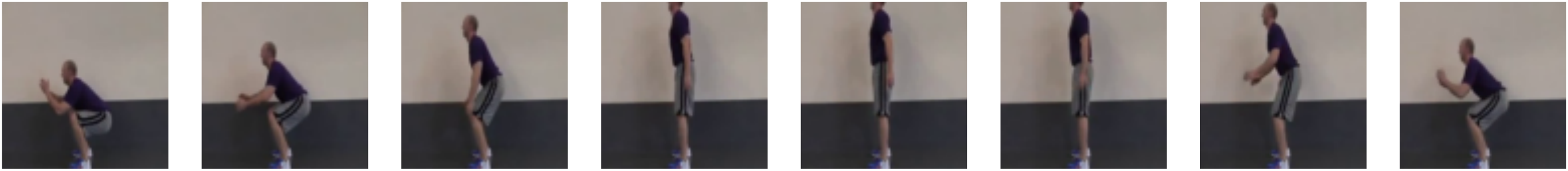}\\
    (a)
    \end{minipage}
    \medskip

    \begin{minipage}{0.9\linewidth}
    \centering
    \includegraphics[width=\linewidth]{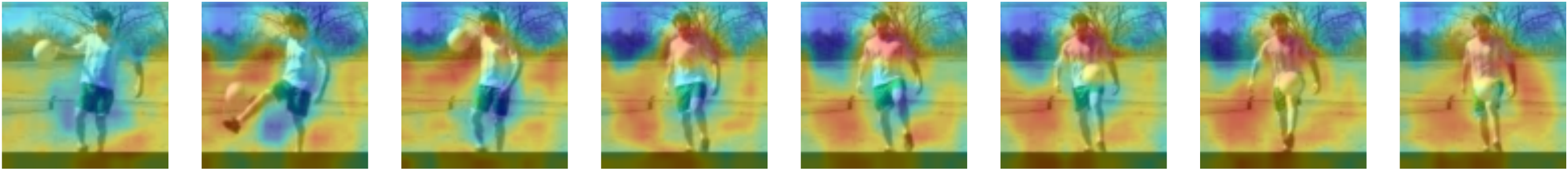}\\
    \includegraphics[width=\linewidth]{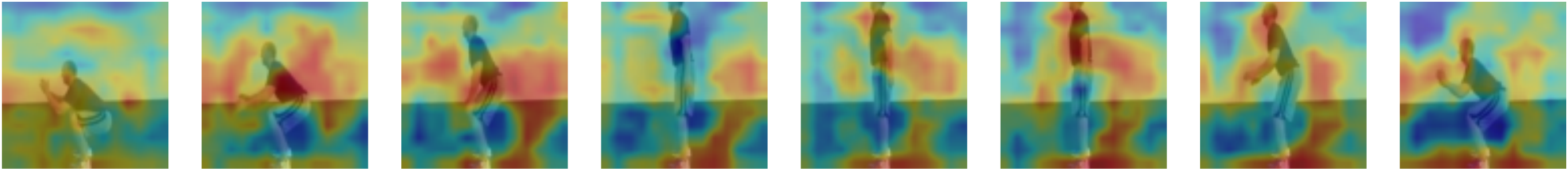}\\
    (b) $M_u$ with $\mathcal{L}_\mathit{per/attn}$
    \end{minipage}
    \medskip

    \begin{minipage}{0.9\linewidth}
    \centering
    \includegraphics[width=\linewidth]{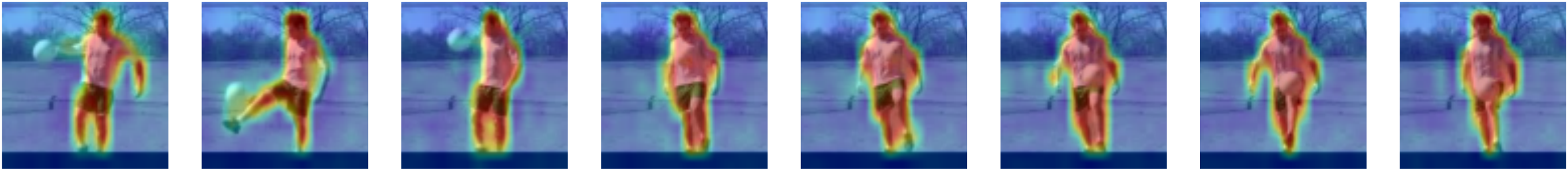}\\
    \includegraphics[width=\linewidth]{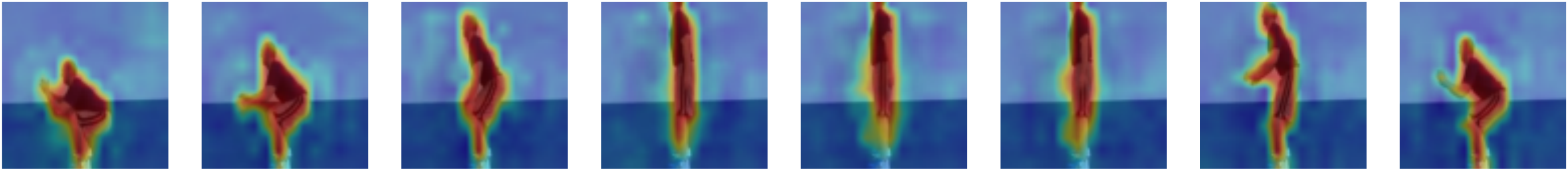}\\
    (c) $M_o$ with $\mathcal{L}_\mathit{per/attn} + \mathcal{L}_\mathit{mask}$
    \end{minipage}
    \medskip
    
    \begin{minipage}{0.9\linewidth}
    \centering
    \includegraphics[width=\linewidth]{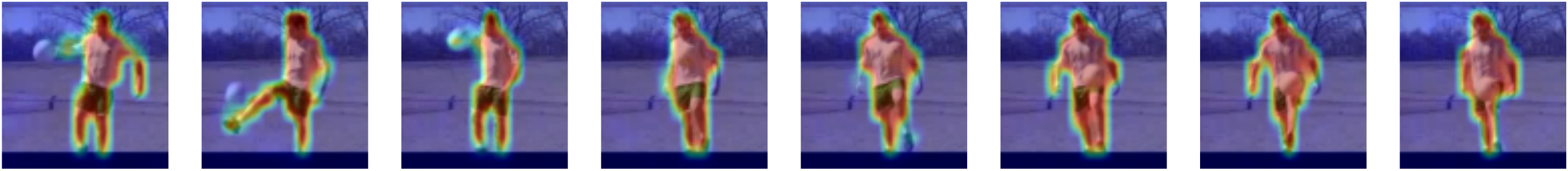}\\
    \includegraphics[width=\linewidth]{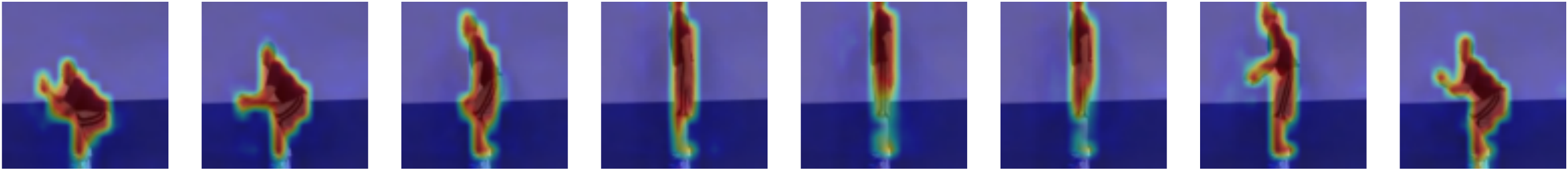}\\
    (d) $M_o$ with $\mathcal{L}_\mathit{per/attn} + \mathcal{L}_\mathit{mask} + \mathcal{L}_\mathit{PC}$
    \end{minipage}

    \caption{Visualization of results for the validation set of UCF101.
    (a) Input videos (every two frames out of 16 frames of a clip are shown).
    (b) Unconstrained maps $M_u$ with $\mathcal{L}_\mathit{per/attn}$ (original ABN).
    (c) Object-constrained maps $M_o$ with $\mathcal{L}_\mathit{per/attn}$, and $\mathcal{L}_\mathit{mask}$, as well as 
    (d) $\mathcal{L}_\mathit{PC}$.
    }

    \label{fig:ABN_results}
\end{figure}

\begin{figure}[t]
    \centering

    \begin{minipage}{0.9\linewidth}
    \centering
    \includegraphics[width=\linewidth]{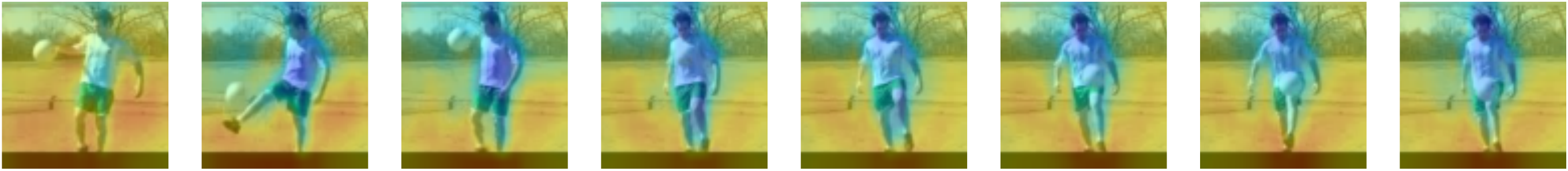}\\
    \includegraphics[width=\linewidth]{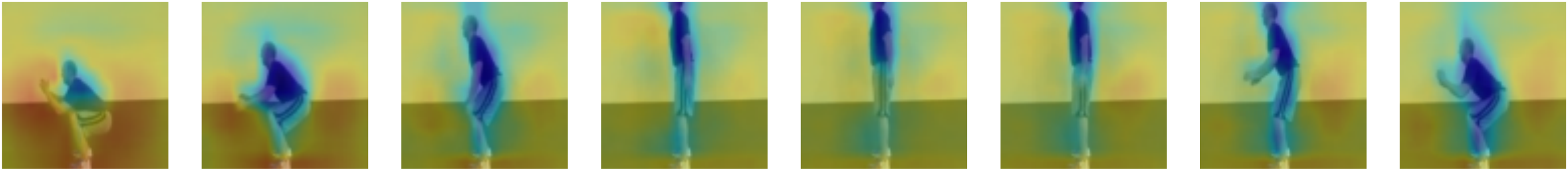}\\
    (a) $M_u$ with $\mathcal{L}_\mathit{per/attn}$ and $\mathcal{L}_\mathit{mha}$
    \end{minipage}
    \medskip

    \begin{minipage}{0.9\linewidth}
    \centering
    \includegraphics[width=\linewidth]{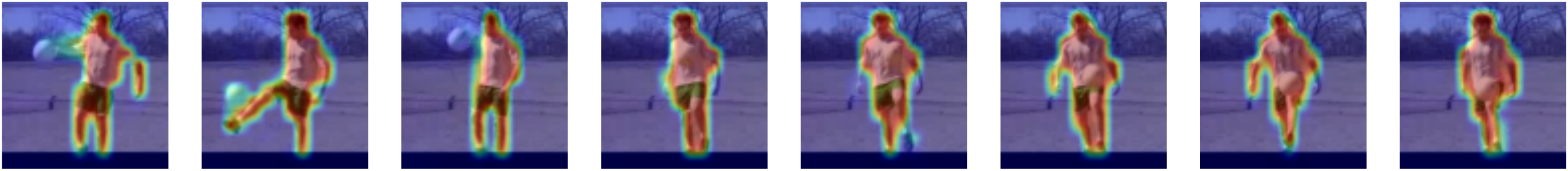}\\
    \includegraphics[width=\linewidth]{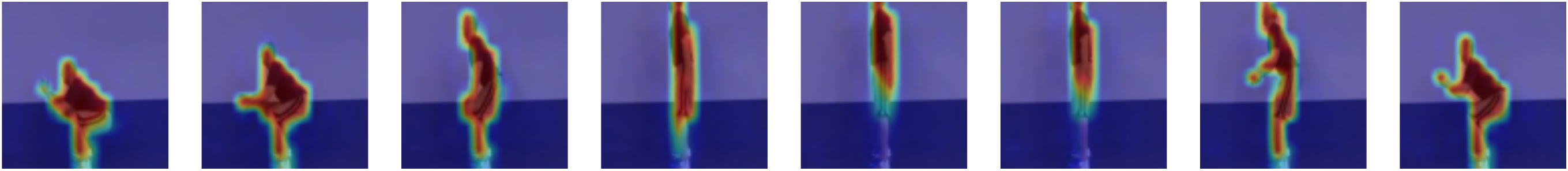}\\
    (b) $M_o$ with $\mathcal{L}_\mathit{per/attn}$ and $\mathcal{L}_\mathit{mha}$
    \end{minipage}
    \medskip

    \begin{minipage}{0.9\linewidth}
    \centering
    \includegraphics[width=\linewidth]{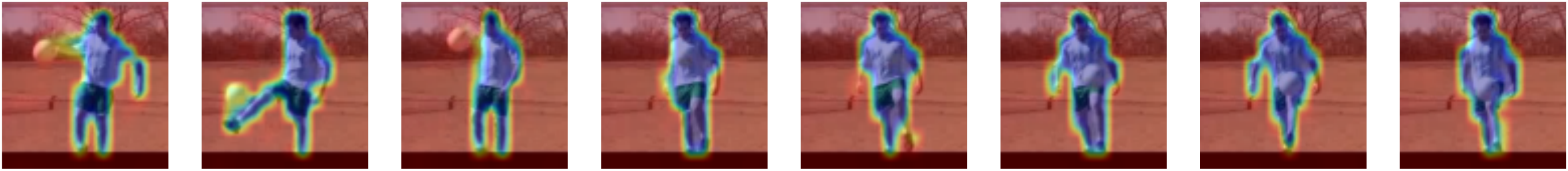}\\
    \includegraphics[width=\linewidth]{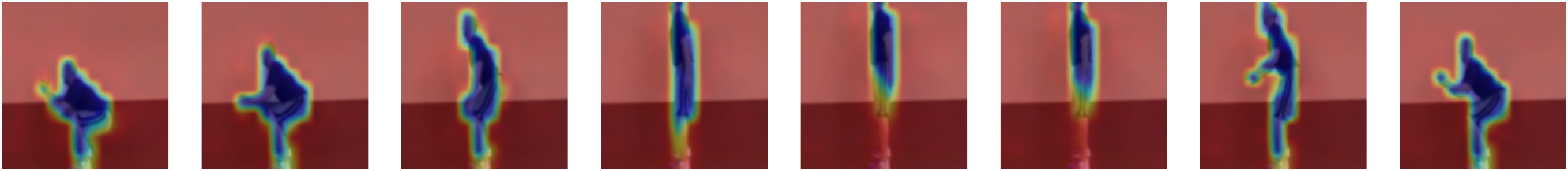}\\
    (c) $M_b$ with $\mathcal{L}_\mathit{per/attn}$ and $\mathcal{L}_\mathit{mha}$
    \end{minipage}

    \caption{
    Visualization of results for the validation set of UCF101 with
    $\mathcal{L}_\mathit{per/attn}$ and $\mathcal{L}_\mathit{mha}$.
    Three types of maps 
    (a) $M_u$,
    (b) $M_o$, and 
    (c) $M_b$.
    }
    \label{fig:multi}
    
\end{figure}

\subsection{Experimental results for UCF101}

We obtained main results with UCF101, which are shown in Tab.\ref{tab:1}.
Each row shows the performance and entropy with different configurations of losses.
The first row with $\mathcal{L}_\mathit{per/attn}$ only is equivalent to the original ABN,
and the second rows is the original ABN with the PC loss $\mathcal{L}_\mathit{PC}$.
In the following rows, results are of Object-ABN
with either of $\mathcal{L}_\mathit{mask}$ and $\mathcal{L}_\mathit{mha}$ is used,
and with or without the PC loss.

\subsubsection{Mask loss}

First, we compare the original ABN
with the proposed Object-ABN to verify the effect of the mask loss.
As can be seen from the first row ($\mathcal{L}_\mathit{per/attn}$ only)
and third row ($\mathcal{L}_\mathit{per/attn}$ and $\mathcal{L}_\mathit{mask}$) of Tab.\ref{tab:1},
the difference in the top-1 performance is 0.3 and not so large.
However, the entropy decreased by more than 1 when the mask loss is used,
which means that quantitatively
the sharpness of the attention map was drastically improved.

Also, the generated attention maps are completely different.
In case of using the mask loss (Fig.\ref{fig:ABN_results}(c)),
generated maps $M_o$ are sharp so that objects and people are clearly visible,
while the case without the mask loss (Fig.\ref{fig:ABN_results}(b))
produced maps $M_u$ that are blurry and speckled,
and action-related foreground and background doesn't appear.

\subsubsection{PC Loss}

Next, we see how the PC loss affect the performance and the attention maps.
As shown in Tab.\ref{tab:1},
using the PC loss reduces the entropy values and improve the performances 
for the cases with the original ABN (the first two rows)
and Object-ABN with $\mathcal{L}_\mathit{mask}$ (the two middle rows).
However, 
when $\mathcal{L}_\mathit{mha}$ is used,
the PC loss seems not to contribute the improvements.

Figures \ref{fig:ABN_results}(d) shows maps with the PC loss.
Compared with Fig.\ref{fig:ABN_results}(c) without the PC loss,
the separation of the background and foreground is clearer,
and the fluctuations in the background disappear.

\subsubsection{Multiple attention maps}

Here, we shows the effect of multi-head attention maps.
Corresponding results are the last two rows of Tab.\ref{tab:1},
where the performance dropped by about 5 \% compared to the cases without $\mathcal{L}_\mathit{mha}$,
even with the PC loss.
Hence, in terms of performance, using a single attention map would be better.

However 
the entropy values becomes smaller and
the quality of the maps were further improved 
by using $\mathcal{L}_\mathit{mha}$,
as shown in Fig.\ref{fig:multi}.
The object-constrained mask $M_o$ (Fig.\ref{fig:multi}(b)) are
far more sharper than those with $\mathcal{L}_\mathit{mask}$ (Fig.\ref{fig:ABN_results}(c)),
and even with $\mathcal{L}_\mathit{PC}$ (Fig.\ref{fig:ABN_results}(d)).

\subsubsection{Comparisons with other methods}

Table \ref{tab:1} also shows performances of other methods that 
are intrinsic models for action recognition for the purpose of visualizing attention maps.
Of course none of them have published entropy values, however
their results are visually much worse than our results in terms of the sharpens of the attention maps.
The performances of our method are comparable or better,

\begin{figure}[t]
    \centering

    \begin{minipage}{0.9\linewidth}
    \centering
    \includegraphics[width=\linewidth]{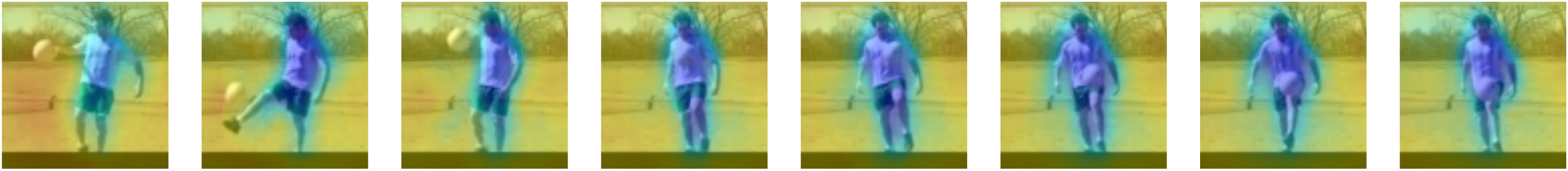}\\
    \includegraphics[width=\linewidth]{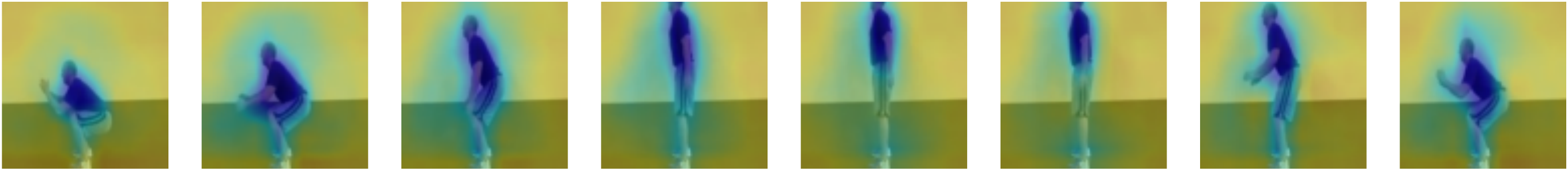}\\
    (a) $M_u$ with $\mathcal{L}_\mathit{per/attn} + \mathcal{L}_\mathit{mha} + \mathcal{L}_\mathit{PC}$
    \end{minipage}
    \medskip

    \begin{minipage}{0.9\linewidth}
    \centering
    \includegraphics[width=\linewidth]{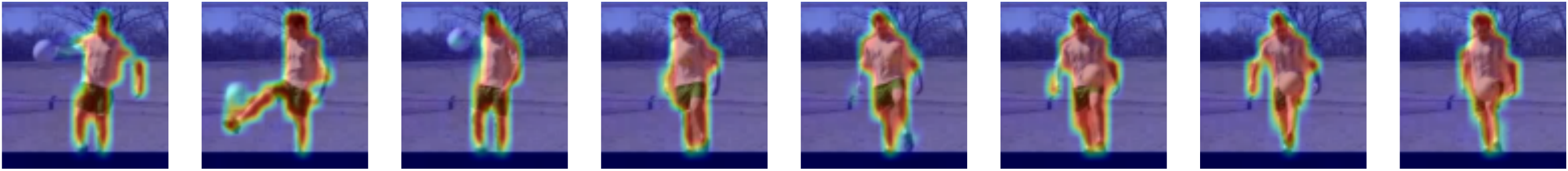}\\
    \includegraphics[width=\linewidth]{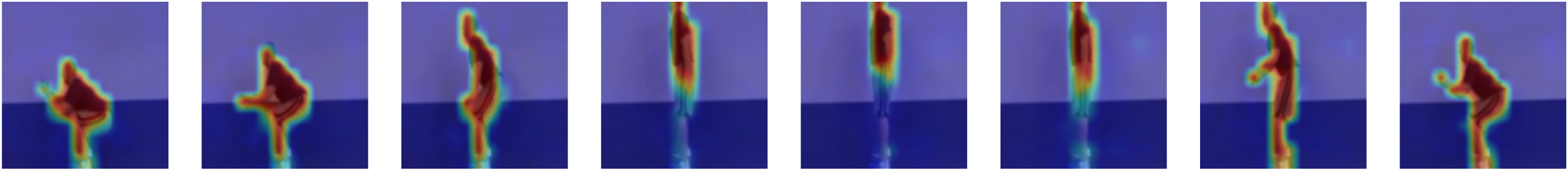}\\
    (b) $M_o$ with $\mathcal{L}_\mathit{per/attn} + \mathcal{L}_\mathit{mha} + \mathcal{L}_\mathit{PC}$
    \end{minipage}
    \medskip

    \begin{minipage}{0.9\linewidth}
    \centering
    \includegraphics[width=\linewidth]{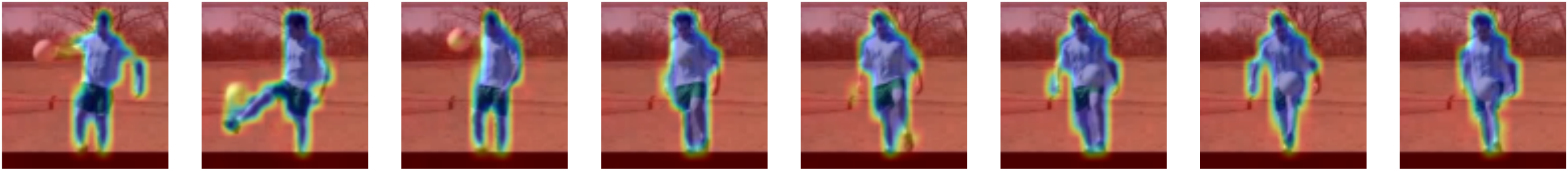}\\
    \includegraphics[width=\linewidth]{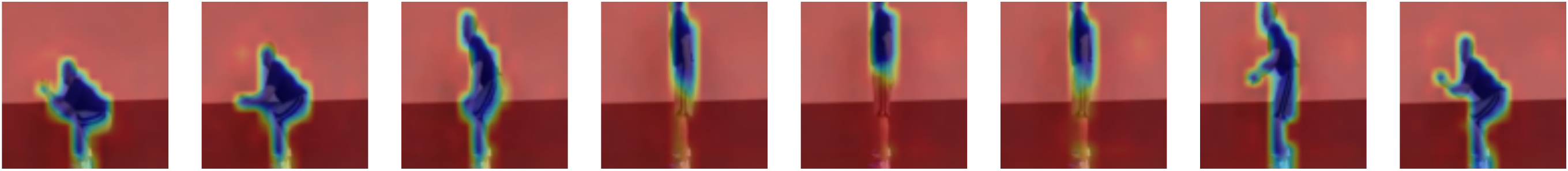}\\
    (c) $M_b$ with $\mathcal{L}_\mathit{per/attn} + \mathcal{L}_\mathit{mha} + \mathcal{L}_\mathit{PC}$
    \end{minipage}

    \caption{
    Visualization of results for the validation set of UCF101 with
    $\mathcal{L}_\mathit{per/attn}, \mathcal{L}_\mathit{mha}$, and $\mathcal{L}_\mathit{PC}$.
    Three types of maps 
    (a) $M_u$,
    (b) $M_o$, and 
    (c) $M_b$.
    }
    \label{fig:multi_pcl}
\end{figure}

\subsection{Experimental result for SSv2}

\begin{figure}[t]
    \centering

    \begin{minipage}{0.9\linewidth}
    \centering
    \includegraphics[width=\linewidth]{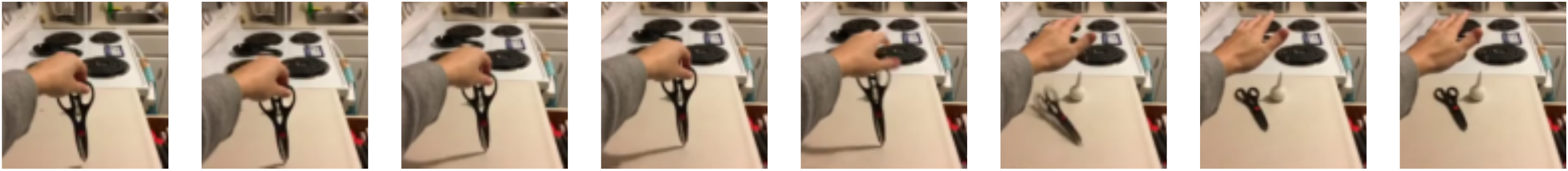}\\
    \includegraphics[width=\linewidth]{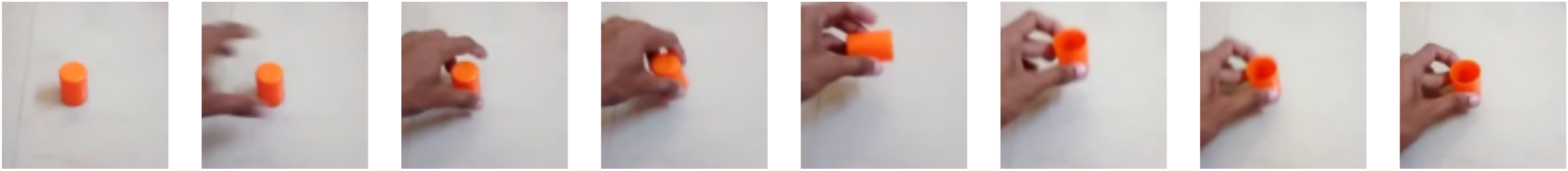}\\
    (a)
    \end{minipage}
    \medskip

    \begin{minipage}{0.9\linewidth}
    \centering
    \includegraphics[width=\linewidth]{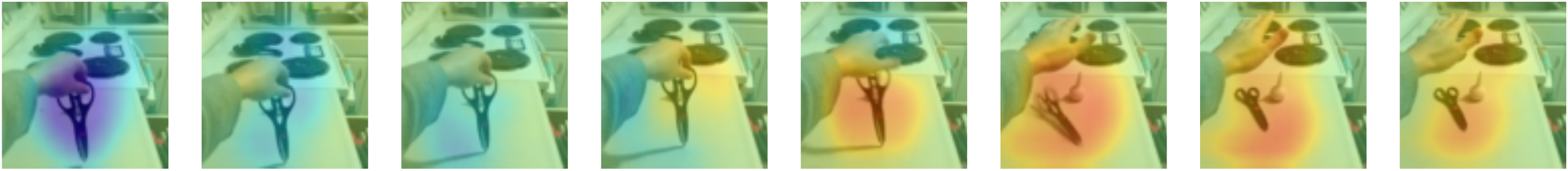}\\
    \includegraphics[width=\linewidth]{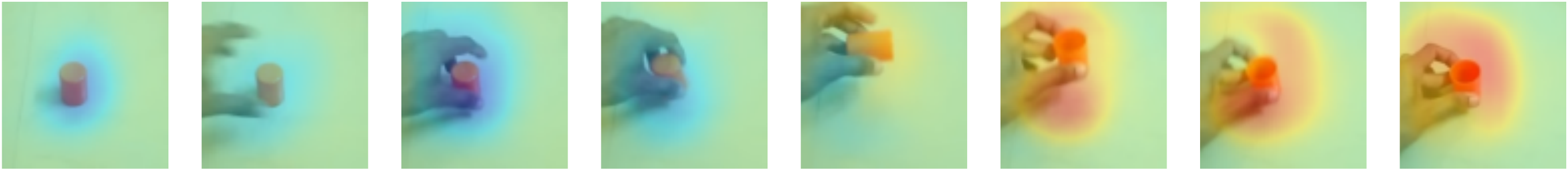}\\
    (b) $M_u$ with $\mathcal{L}_\mathit{per/attn}$
    \end{minipage}
    \medskip
    
    \begin{minipage}{0.9\linewidth}
    \centering
    \includegraphics[width=\linewidth]{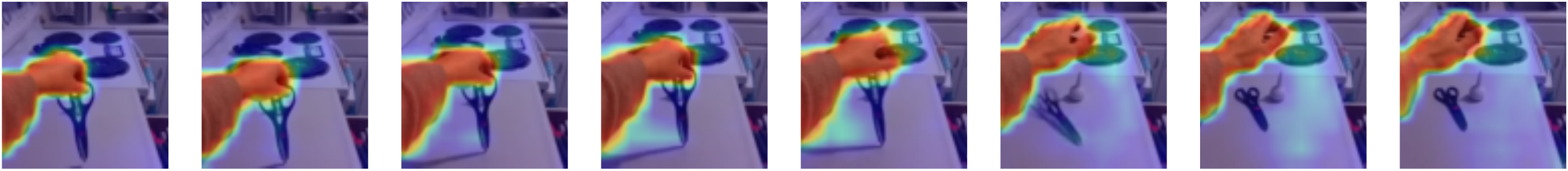}\\
    \includegraphics[width=\linewidth]{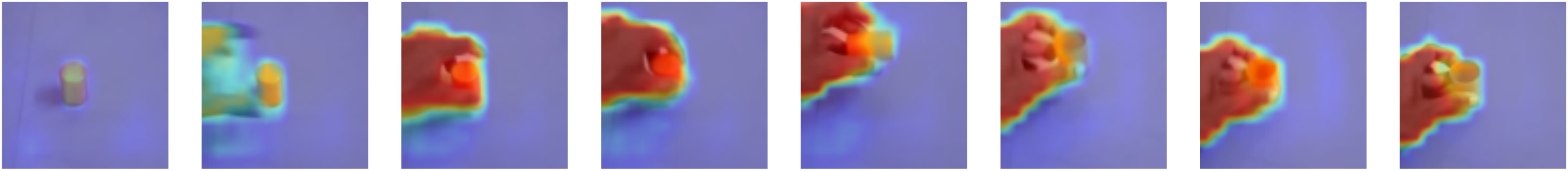}\\
    (c) $M_o$ with $\mathcal{L}_\mathit{per/attn} + \mathcal{L}_\mathit{mask}$
    \end{minipage}
    \medskip
    
    \begin{minipage}{0.9\linewidth}
    \centering
    \includegraphics[width=\linewidth]{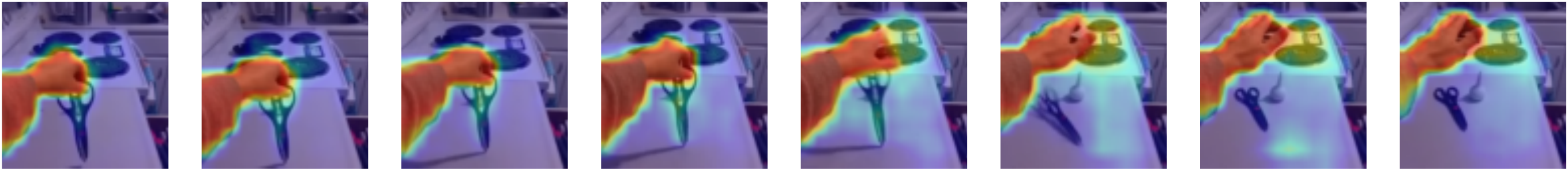}\\
    \includegraphics[width=\linewidth]{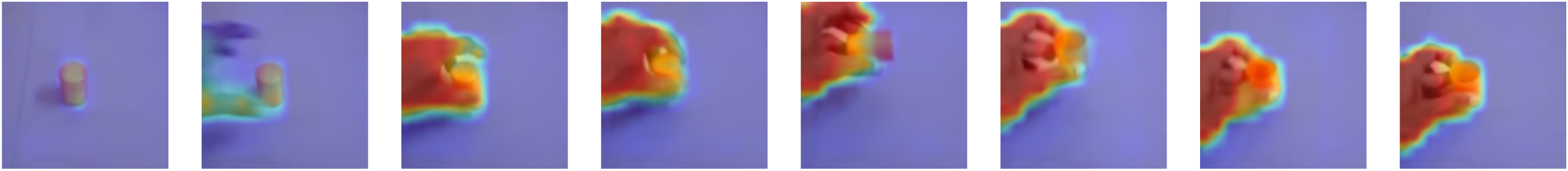}\\
    (d) $M_o$ with $\mathcal{L}_\mathit{per/attn} + \mathcal{L}_\mathit{mask} + \mathcal{L}_\mathit{PC}$
    \end{minipage}
    
    \caption{Visualization of results for the validation set of SSv2.
    (a) Input videos.
    (b) $M_u$ with $\mathcal{L}_\mathit{per/attn}$.
    (c) $M_o$ with $\mathcal{L}_\mathit{per/attn}$, and $\mathcal{L}_\mathit{mask}$, as well as 
    (d) $\mathcal{L}_\mathit{PC}$.
    }

    \label{fig:ssv2}
\end{figure}

\begin{figure}[t]
    \centering

    \begin{minipage}{0.9\linewidth}
    \centering
    \includegraphics[width=\linewidth]{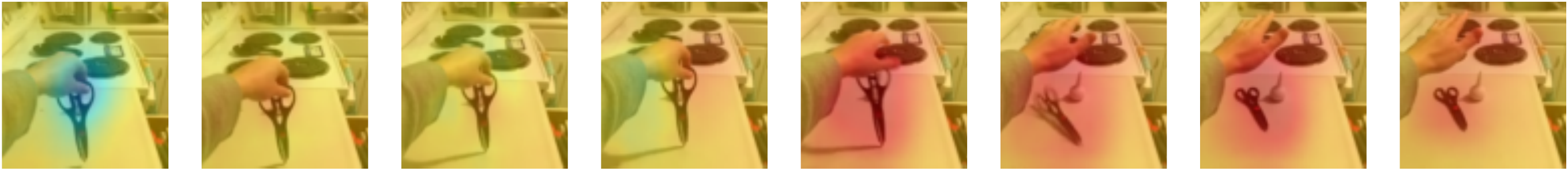}\\
    \includegraphics[width=\linewidth]{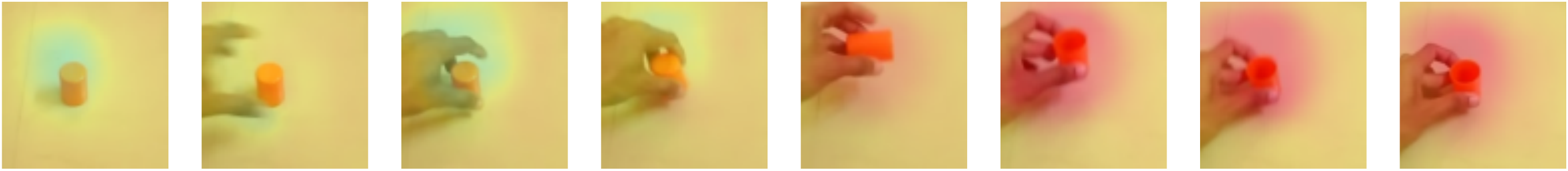}\\
    (a) $M_u$ with $\mathcal{L}_\mathit{per/attn} + \mathcal{L}_\mathit{mha} + \mathcal{L}_\mathit{PC}$
    \end{minipage}
    \medskip

    \begin{minipage}{0.9\linewidth}
    \centering
    \includegraphics[width=\linewidth]{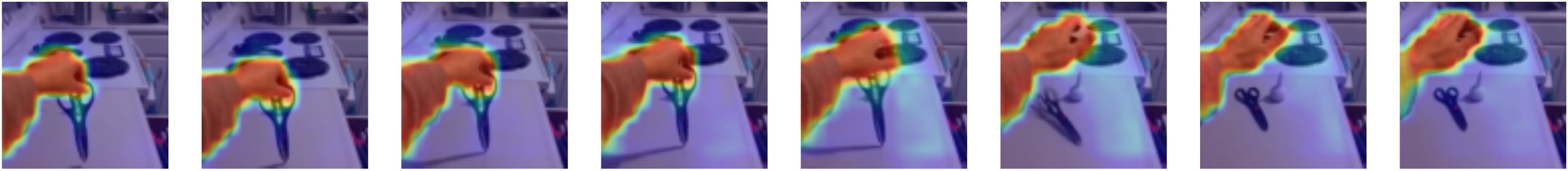}\\
    \includegraphics[width=\linewidth]{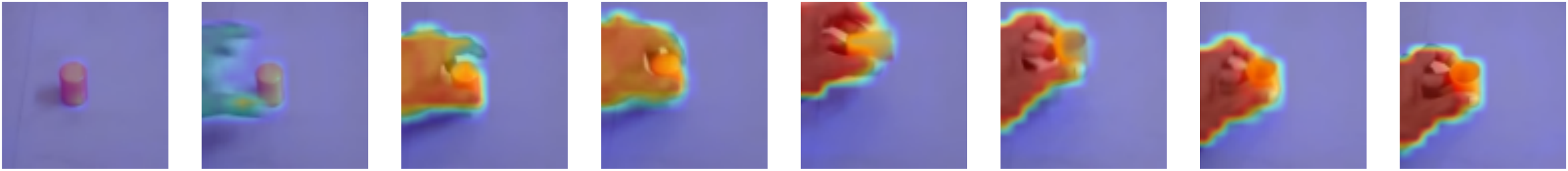}\\
    (b) $M_o$ with $\mathcal{L}_\mathit{per/attn} + \mathcal{L}_\mathit{mha} + \mathcal{L}_\mathit{PC}$
    \end{minipage}
    \medskip

    \begin{minipage}{0.9\linewidth}
    \centering
    \includegraphics[width=\linewidth]{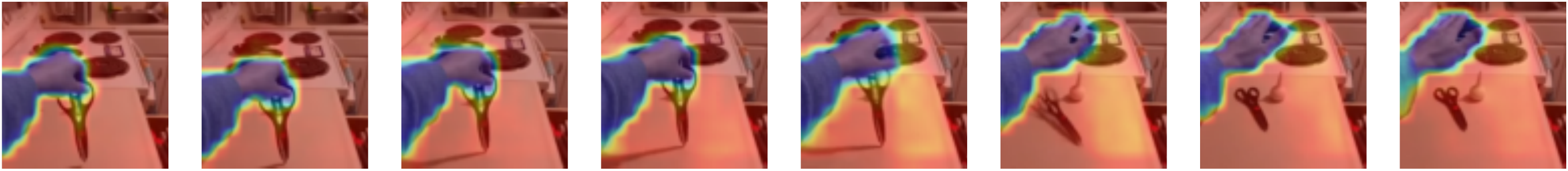}\\
    \includegraphics[width=\linewidth]{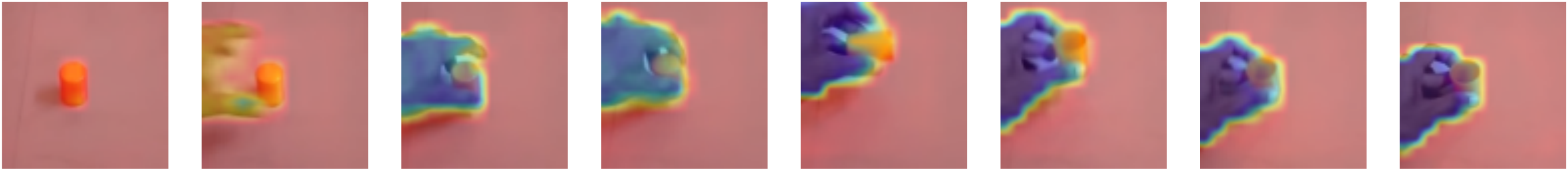}\\
    (c) $M_b$ with $\mathcal{L}_\mathit{per/attn} + \mathcal{L}_\mathit{mha} + \mathcal{L}_\mathit{PC}$
    \end{minipage}
    \hfill

    \caption{
    Visualization of results for the validation set of SSv2 with
    $\mathcal{L}_\mathit{per/attn}, \mathcal{L}_\mathit{mha}$, and $\mathcal{L}_\mathit{PC}$.
    Three types of maps 
    (a) $M_u$,
    (b) $M_o$, and 
    (c) $M_b$.
    }

    \label{fig:ssv2_multi}
\end{figure}

Here we show the experimental results for SSv2.
The training settings was the same with UCF101, except for
the frame sampling stride (two frames instead of four),
no horizontal flip, and 25 epochs for training.
Performances are shown in Tab.\ref{tab:ssv2},
and visualization results with the mask loss are shown in Fig.\ref{fig:ssv2},
and with the multiple attention maps in Fig.\ref{fig:ssv2_multi}.

The performance was improved from the original ABN by adding the mask loss,
and the entropy of $M_o$ for the Object-ABN is smaller than that of $M_u$ for ABN.
Furthermore, using the PC loss and adding the multiple attention maps
improve the performance.
The object-constrained maps $M_o$ shown in Fig.\ref{fig:ssv2}(c)(d)
and Fig.\ref{fig:ssv2_multi}(b) look almost the same,
supported by the similar entropy values in Tab.\ref{tab:ssv2}.
This suggests that for this dataset
the mask loss has the largest impact on the sharpness of the maps,
while the multiple attention maps and the PC loss also contribute to the performance.

The unconstrained maps $M_u$ are shown in
Fig.\ref{fig:ssv2}(b) for ABN
and Fig.\ref{fig:ssv2_multi}(a) for the proposed method.
For the maps of ABN,
the attentions to the object and hands are weaker (in blue) than to the background at the beginning.
After the action starts, the attention is getting focused on the object,
then becomes strong at the end.
In contrast,
the maps of the proposed method are flat at first,
then the attention is continuously focused on the object during the action until the end.
Therefore, the maps of the proposed method are expected to better represent the temporal information of the action.

\begin{table}[t]
    \centering

    \caption{
        The performances and entropy values
        with different configurations for the validation set of SSv2.
    }
    \label{tab:ssv2}

    \scalebox{\tabscaleA}{
        \begin{tabular}{c@{\ }c@{\ }c@{\ }c@{\ }|@{\hspace{.5em}}c@{\hspace{.5em}}c@{\hspace{.5em}}c@{\hspace{.5em}}c@{\hspace{.5em}}c}
            &&&&& & entropy & entropy & entropy\\
             $\mathcal{L}_\mathit{per/attn}$ & $\mathcal{L}_\mathit{mask}$ & $\mathcal{L}_\mathit{mha}$ & 
             $\mathcal{L}_\mathit{PC}$ & top-1 & top-5 & $M_u$ & $M_o$ & $M_b$ \\
             \hline
             \checkmark & & & & 54.63 & 83.23 & 2.887\\
             \checkmark & \checkmark & & & 54.83 & 82.73 &  & 2.331\\
             \checkmark & \checkmark & & \checkmark &   54.97 &  83.15  &       & 2.427  &     \\
             \checkmark & & \checkmark & \checkmark & 55.05 & 83.60 & 2.921 & 2.236 & 2.237\\
        \end{tabular}
        }
\end{table}

\section{Conclusion}

In this paper, we have proposed Object-ABN, an extension of ABN by using instance segmentation,
and enables the generation of sharper attention maps,
which enable us to clearly see which parts of the scene the model is focusing on.
Experiments with two datasets demonstrated that the proposed method with the mask loss,
multiple attention maps, and the PC loss improves the quality of attention maps
in terms of entropy, as well as the classification performances.

{\small
\bibliographystyle{ieee_fullname}
\bibliography{mybib}
}

\end{document}